\documentclass{article}
\usepackage{spconf,amsmath,graphicx}
\usepackage{color}
\usepackage{epsfig}
\usepackage{subfigure}
\usepackage{graphicx}
\usepackage{epstopdf}
\usepackage{caption}
\usepackage{booktabs}

\title{Integrated Deep and Shallow networks for salient object detection}
\name{Jing Zhang $^{1,2}$, Bo Li $^{1}$, Yuchao Dai$^{2}$, Fatih Porikli$^{2}$ and Mingyi He$^{1}$\thanks{This work was done when Jing Zhang was a CSC  joint PhD student between Northwestern Polytechnical University and the Australian National University/NICTA. This work was supported in part by the Australian Research Council grants (DE140100180, DP150104645), and Natural Science Foundation of China grants (61420106007, 61671387).}}
\address{$^{1}$ School of Electronics and Information, Northwestern Polytechnical University, Xi'an, China \\ $^{2}${ Research School of Engineering, Australian National University, Canberra, Australia}.}

\begin{document}
%
\maketitle
\begin{abstract}



Deep convolutional neural network (CNN) based salient object detection methods have achieved state-of-the-art performance and outperform those unsupervised methods with a wide margin. In this paper, we propose to integrate deep and unsupervised saliency for salient object detection under a unified framework. Specifically, our method takes results of unsupervised saliency (Robust Background Detection, RBD) and normalized color images as inputs, and directly learns an end-to-end mapping between inputs and the corresponding saliency maps. The color images are fed into a Fully Convolutional Neural Networks (FCNN) adapted from semantic segmentation to exploit high-level semantic cues for salient object detection. Then the results from deep FCNN and RBD are concatenated to feed into a shallow network to map the concatenated feature maps to saliency maps. Finally, to obtain a spatially consistent saliency map with sharp object boundaries, we fuse superpixel level saliency map at multi-scale. Extensive experimental results on 8 benchmark datasets demonstrate that the proposed method outperforms the state-of-the-art approaches with a margin.


\end{abstract}
\begin{keywords}
Salient object detection, fully convolutional neural networks, shallow network, robust background detection, multi-scale fusion
\end{keywords}
\section{Introduction}
\label{sec:intro}
Salient object detection aims at identifying visually interesting object regions that are consistent with human perception. It is essential in many computer vision tasks including object-aware image retargeting \cite{Wang:image_resizing_2008} interactive image segmentation \cite{Segmentation_saliency} and so on.
\begin{figure}[!htp]
   \begin{center}
   {\includegraphics[width=0.13\linewidth]{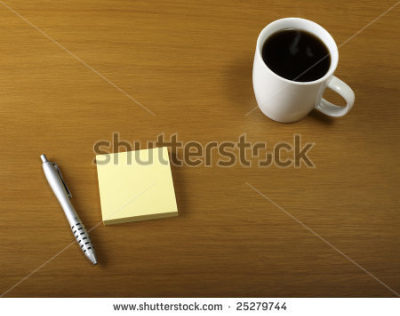}}
   {\includegraphics[width=0.13\linewidth]{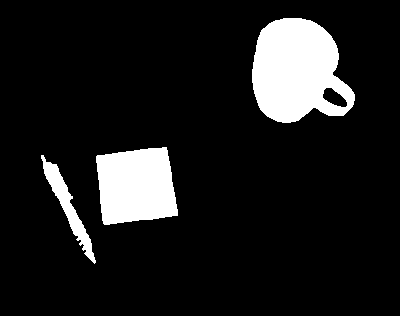}}
   {\includegraphics[width=0.13\linewidth]{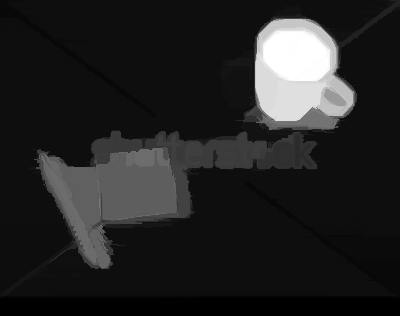}}
   {\includegraphics[width=0.13\linewidth]{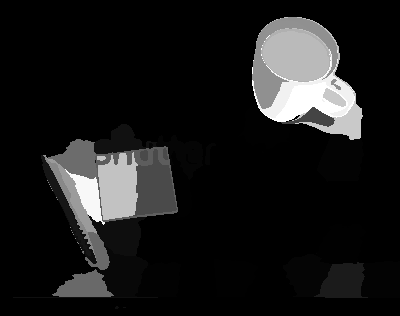}}
   {\includegraphics[width=0.13\linewidth]{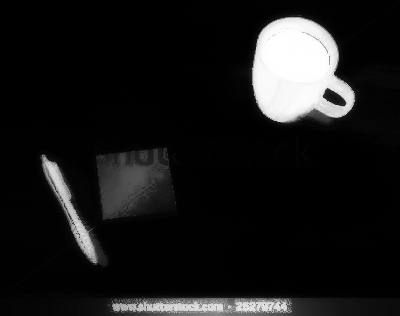}}
   {\includegraphics[width=0.13\linewidth]{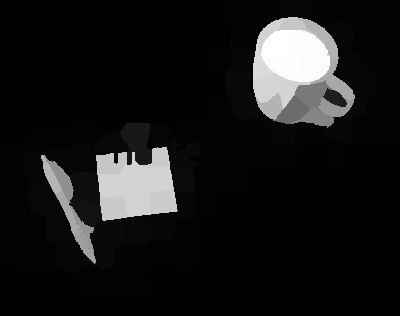}}
   {\includegraphics[width=0.13\linewidth]{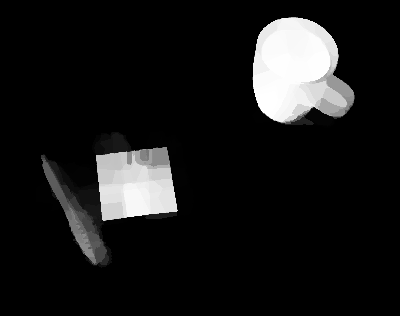}}
   {\includegraphics[width=0.13\linewidth]{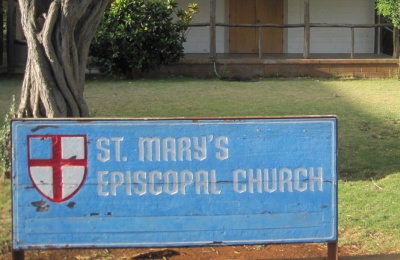}}
   {\includegraphics[width=0.13\linewidth]{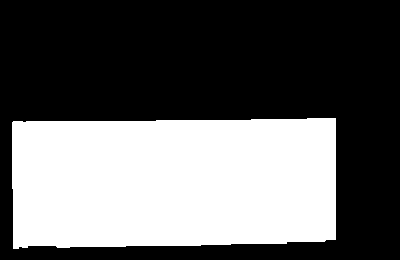}}
   {\includegraphics[width=0.13\linewidth]{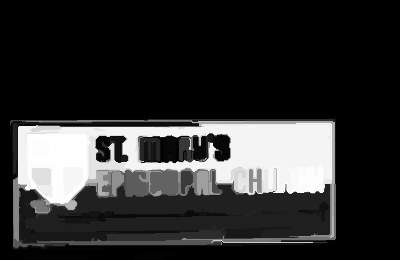}}
   {\includegraphics[width=0.13\linewidth]{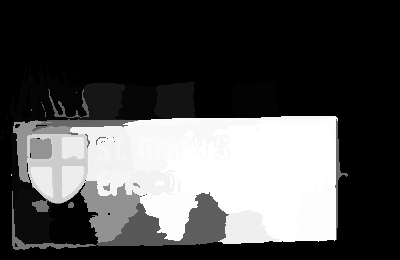}}
   {\includegraphics[width=0.13\linewidth]{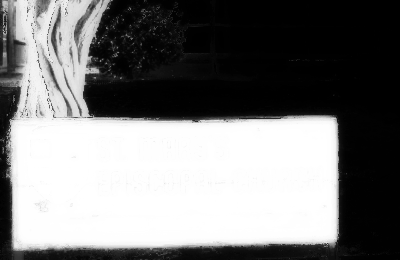}}
   {\includegraphics[width=0.13\linewidth]{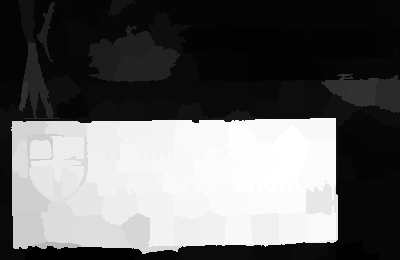}}
   {\includegraphics[width=0.13\linewidth]{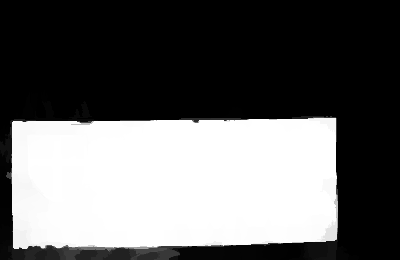}}
   \end{center}
   \vspace{-5mm}
   \caption{Saliency detection results where unsupervised method (RBD) outperforms state-of-the-art deep learning based methods. From left to right: Input image, ground truth, result of MDF \cite{MDF:CVPR-2015}, DeepMC \cite{DeepMC}, DC \cite{DC}, RBD \cite{Background-Detection:CVPR-2014} and our method.}
   \label{fig:saliency_compare1}
\end{figure}
Most of the traditional saliency detection methods are based on low-level hand-crafted features such as color and texture descriptors, or they compute variants of appearance uniqueness and region compactness based on statistical priors, e.g. center prior \cite{ContextSaliency} and boundary prior \cite{Background-Detection:CVPR-2014}. These methods report acceptable results on relatively simple datasets, but their saliency maps deteriorate when the input images become cluttered and complicated.

Recently, deep learning has achieved great success in high-level computer vision tasks, and many deep learning based saliency detection methods \cite{TIP} \cite{MDF:CVPR-2015} \cite{DeepMC} \cite{DC} \cite{RFCN} \cite{DISC} have been proposed to learn competent high-level feature representations for salient objects, which achieve state-of-the-art performance and outperform unsupervised saliency detection methods. These methods are either built on semantic segmentation \cite{DC} (to leverage high-level semantic cues) or they learn saliency features by exploiting different datasets \cite{TIP}.

\begin{figure*}[!htp]
   \begin{center}
   {\includegraphics[width=0.82\linewidth]{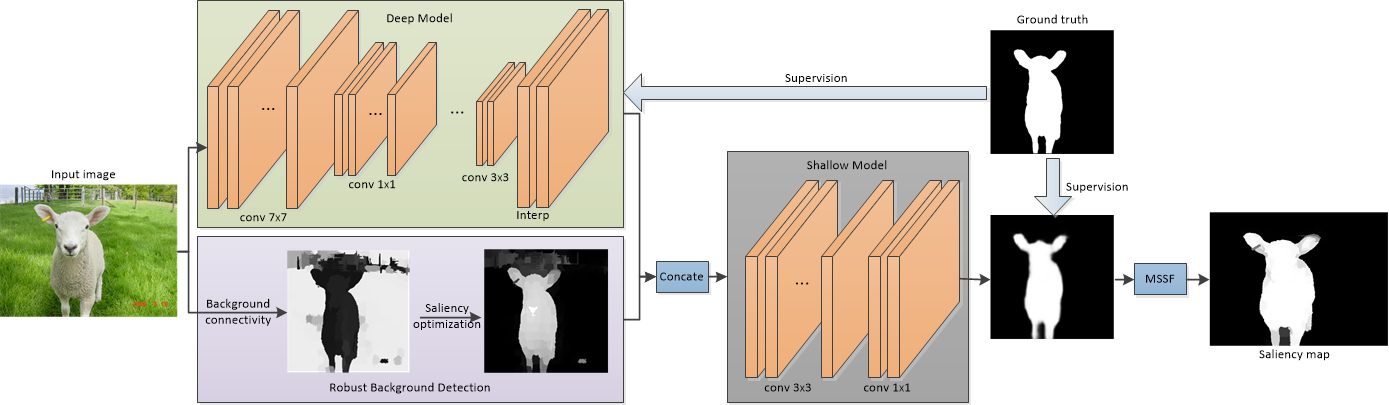}}
   \end{center}
   \vspace{-5mm}
   \caption{A nutshell of our framework. Given a color image, the deep model outputs a coarse saliency map, which is concatenated with unsupervised saliency map from RBD to feed to our shallow model. Multi-scale superpixel level fusion (MSSF) is adopted to obtain a spatially coherent saliency map. We apply the supervision to both intermedia and final outputs in the network.}
   \label{fig:Nutshell_Overview}
\end{figure*}

The basic priors (center prior, boundary prior and etc.) used in unsupervised saliency detection methods are summarized and described with human knowledge. They are more universal and applicable to general cases. Even though the performance has been outperformed by deep learning based methods, there still exit scenarios where unsupervised methods outperform deep learning based methods, see Fig.~\ref{fig:saliency_compare1} for examples. This naturally raises a question that whether the data-driven deep learning based saliency detection methods have sufficiently exploited the statistics of saliency. In this paper, we investigate the problem that ``Could unsupervised saliency and deep saliency benefit each other to achieve even better performance?'' A positive answer will provide deeper insight to understand the nature of salient object detection.



To this end, we propose to bridge the deep supervised saliency with unsupervised saliency by integrating deep and unsupervised saliency in a unified framework for salient object detection. Our method takes results of unsupervised saliency (RBD \cite{Background-Detection:CVPR-2014}) and normalized color images as inputs, and directly learns an end-to-end mapping between inputs and the corresponding saliency maps. Then multi-scale saliency fusion is performed to get a spatially consistent saliency map.

\section{Our method}
\subsection{Overview}
Targeting at exploiting how deep saliency and unsupervised saliency can be complementary to each other, we propose a deep fully convolutional neural network to integrate both deep and unsupervised saliency for salient object detection.


First, we reconstruct a Convolutional Neural Network (CNN) \cite{ResHe2015} to adapt it to salient object detection (which is our deep model) to learn a coarse level per-pixel saliency map. Then a shallow model is used to take concatenation of the deep saliency map and result of RBD as input, and learns our deep-shallow saliency map. Input of our model includes two parts: normalized RGB image $I$ (resized to $224\times 224\times 3$) and saliency map using exiting unsupervised method RBD $\rm S_{RBD}$ \cite{Background-Detection:CVPR-2014} (resized to $224\times 224$). $I$ is fed to our deep model to get a two channel feature map which is 1/8 size of the normalized RGB image. Then an interpolation layer is adopted to upsample the feature map to $224\times 224$, and a loss layer is used to transfer this two channel feature maps to deep saliency map $\rm S_{deep}$. Furthermore, we concatenate $\rm S_{deep}$ and $\rm S_{RBD}$, and feed them to our shallow model to get a deep-shallow integrated saliency map $\rm S_{DS}$ which is also $224\times 224$. The above models are trained in deep supervised manner, where both the deep model and the shallow model are trained simultaneously. We end up with a coarse and dense saliency map with blurred object boundaries. Finally, multi-scale superpixel level saliency fusion (MSSF) is adopt to get a spatially coherence saliency map $\rm S_{DSM}$. A nutshell of our framework is illustrated in Fig.\ref{fig:Nutshell_Overview}.


\subsection{FCNN based Deep Saliency Detection}

Our salient object detection network (deep model) is built upon the DeepLab semantic segmentation network \cite{Deeplab}, where a deep convolutional neural network (ResNet-101 \cite{ResHe2015}) originally designed for image classification is re-purposed to the task of semantic segmentation by 1) transforming all the fully connected layers to convolutional layers and 2) increasing feature resolution through atrous or dilation convolutional layers \cite{Deeplab}. In this way, the spatial resolution of the output feature map has been increased four times, which is much denser than \cite{DeepMC} \cite{MDF:CVPR-2015}. Different from \cite{Deeplab}, we interpolate the two channel feature map to input image size ($224\times 224$) before the loss layer of our deep model, and compute loss and accuracy using the interpolated feature map. Note that, this upsampling operation could also be achieved by deconvolution, which leads to much more parameters but similar performance. In this paper, we use the ``Interp'' layer provided in Caffe \cite{jia2014caffe} due to its efficiency.

Different from existing deep network \cite{VGG}, Deep Residual Network \cite{ResHe2015} explicitly learns residual functions with reference to the layer inputs, which makes it easier to optimize with higher accuracy from considerably increased depth. By removing the final pooling and fully-connected layer to adapt it for saliency detection, we reconstruct ResNet-101 model to transfer it to Fully Convolutional Neural Networks, and add four dilation convolutional layers with increasing receptive field to better utilize local and global information. Finally a loss layer is used to map the intermedia feature map to a two channel feature map $\rm S_{deep}$ with each channel represents the possibility for each pixel to be background or salient object.

\subsection{Integrating Deep and Unsupervised Saliency}
To further explore the statistics of unsupervised saliency, we integrate deep and unsupervised saliency in a unified framework for salient object detection.
\begin{figure}[!htp]
   \begin{center}
   {\includegraphics[width=0.18\linewidth]{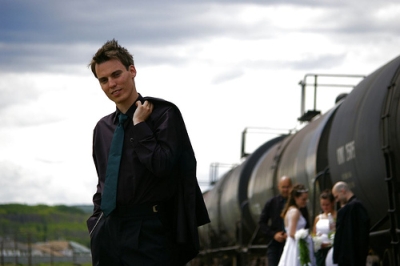}}
   {\includegraphics[width=0.18\linewidth]{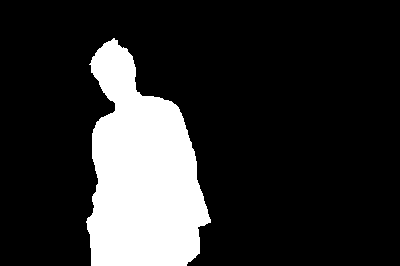}}
   {\includegraphics[width=0.18\linewidth]{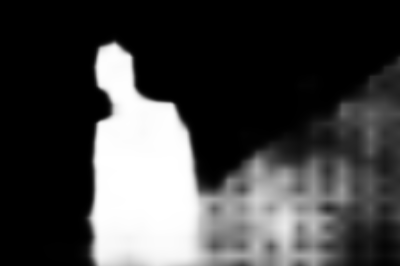}}
   {\includegraphics[width=0.18\linewidth]{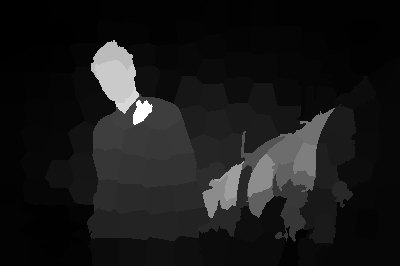}}
   {\includegraphics[width=0.18\linewidth]{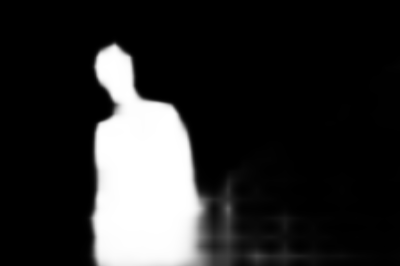}}
   {\includegraphics[width=0.18\linewidth]{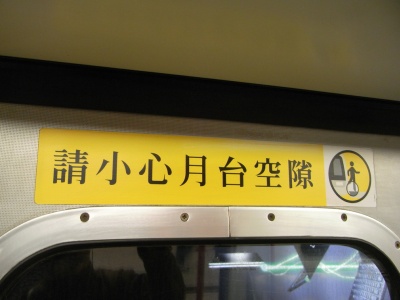}}
   {\includegraphics[width=0.18\linewidth]{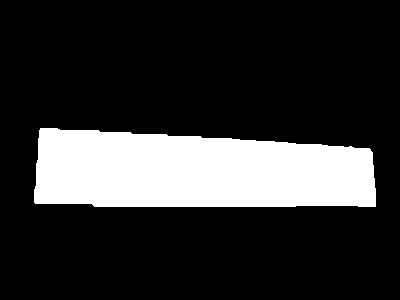}}
   {\includegraphics[width=0.18\linewidth]{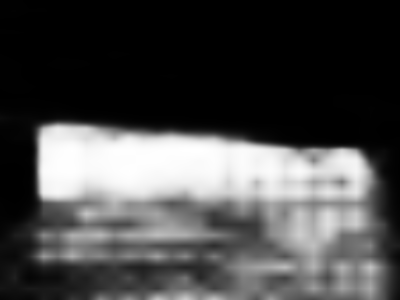}}
   {\includegraphics[width=0.18\linewidth]{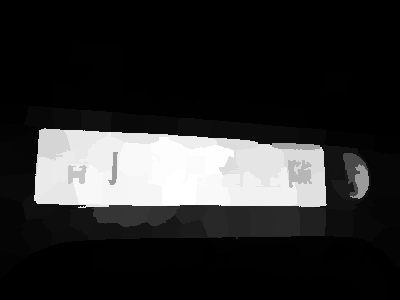}}
   {\includegraphics[width=0.18\linewidth]{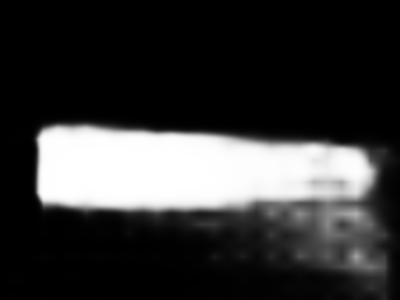}}
   \end{center}
   \vspace{-5mm}
   \caption{Benefit of integrating deep and unsupervised saliency. From left to right: Input image, ground truth, results of deep model, RBD and our deep-shallow integrated model.}
   \label{how_shallow_helps}
\end{figure}RBD ranks 1st of all the unsupervised saliency detection method \cite{SalObjBenchmark_Tip2015}. According to \cite{Background-Detection:CVPR-2014}, background connectivity is defined as: $ BndCon(p)=\frac{Len_{bnd}(p)}{\sqrt[]{Area(p)}}$ (where $Len_{bnd}(p)$ and $Area(p)$ are length along image boundary and spanning area of superpixel $p$ respectively), which provides strong prior in distinguishing salient object from background. In this paper, we take $\rm S_{deep}$ and $\rm S_{RBD}$ as inputs, and construct another shallow network to get our deep-shallow integrated saliency map $\rm S_{DS}$. Firstly, $\rm S_{deep}$ and $\rm S_{RBD}$ are concatenated in channel dimension. Then two convolution layers (our shallow model) are utilized to map the concatenated feature maps to another 2 channel feature maps with a loss layer in the end.

To illustrate how unsupervised saliency benefits deep saliency, we show results of $\rm S_{deep}$ and $\rm S_{DS}$ in Fig.~\ref{how_shallow_helps}, where $\rm S_{deep}$ is trained without $\rm S_{RBD}$ as input. Fig.~\ref{how_shallow_helps} illustrates that with the help of RBD (2nd row), we end up with better saliency map with most of the background suppressed. Furthermore, for situations where RBD fails to capture saliency region (1st row), our model still achieves good result, which proves that deep saliency and unsupervised saliency are complementary under our unified framework.

\subsection{Multi-scale superpixel level saliency map fusion}

With our deep and shallow model, a coarse saliency map is obtained with blurred boundaries. To get a spatially consistent saliency map with sharp object boundaries, multi-scale superpixel level saliency map fusion is performed. For a given image $I$, SLIC \cite{SLIC:PAMI-2012} is used for image over-segmentation to represent $I$ as a collection of superpixels $I=\left\{I_1, I_2, \cdots, I_N\right\}$, where the numbers of superpixels $N=\left\{100,200,300,400\right\}$ are used to achieve multi-scale over-segmentation, which has been widely used in achieving higher resolution saliency prediction map \cite{DRFI:CVPR-2013}. Given saliency detection map $\rm S_{DS}$ from deep and shallow model as above, for a specific number of $N$, we get a saliency score vector $ S_v=\left\{s_{1v},s_{2v},...,s_{Nv}\right\}$ and per-superpixel saliency map $ S_k, k = {1,2,3,4}$, where the per-superpixel score $s_{iv}$ is defined as the median saliency score of superpixel $I_i$. Then our final saliency map $\rm S_{DSM}$ with MSSF refinement is the sum of $ S_k$, i.e., $\rm S_{DSM} = \sum \emph{$S_k$} $. Here we use equal weight for each scale of image. Alternatively, we have constructed another convolutional network to learn the weights for each scale and similar weights have obtained. For efficiency, we use equal weights in our work.


\begin{table*}[!htp] \footnotesize
\begin{center}
\caption{MAE for different methods including ours on eight benchmark datasets.(Best ones in bold)} \centering
\begin{tabular}{l| c| c| c| c| c| c| c| c| c| c| c| c}\hline
  & DC & MDF & DeepMC & DMT & DISC & RFCN & DRFI & RBD & DSR & MC & DS & DSM\\ \hline
ECSSD  & 0.0906 & 0.1081 &  0.1019& 0.1601 & 0.1122 & 0.0973 & 0.1719 & 0.1739 & 0.1742 & 0.2037 &0.0628& \textbf{0.0610} \\ \hline
DUT  & 0.0971 &  0.0916& 0.0885 & 0.0758 & 0.1182 & 0.0945 &0.1496  & 0.1467 & 0.1374 & 0.1863 & 0.0663& \textbf{0.0648}\\ \hline
SED1  &  0.0886&  0.1198 & 0.0881 & - &  0.0772& 0.1020 & 0.1454 & 0.1407 & 0.1614 & 0.1620 & 0.0552 &\textbf{0.0533} \\ \hline
SED2  & 0.1014 & 0.1171 &  0.1162 & 0.1074 & 0.1203 & 0.1140 & 0.1373 & 0.1316 & 0.1457 & 0.1848 & 0.0876 &\textbf{0.0802}\\ \hline
PASCALS  &  0.1246 & 0.1420 &  0.1422& 0.1695 &-  & 0.1176 & 0.2071 &0.1985  & 0.2041 & 0.2296 & 0.0902 &\textbf{0.0810}\\ \hline
SOD  &  0.1208&  0.1669& 0.1557 & 0.1503 & - & 0.1394 & 0.2046 & 0.2069 & 0.2133 &  0.2435 & 0.1104 &\textbf{0.1002}\\ \hline
HKU-IS  & 0.0730 & - & 0.0913 & - & 0.1023 & 0.0798 & 0.1445 &  0.1432& 0.1404 & 0.1840 &0.0515 &\textbf{0.0486} \\ \hline
THUR  & 0.0959 & 0.1029 & 0.1025 & 0.0854 & - &0.1003  & 0.1471 &  0.1507&  0.1408&  0.1838& 0.0752 &\textbf{0.0704}\\ \hline
\end{tabular}
\vspace{-5mm}
\label{tab:Performance_Comparison}
\end{center}
\end{table*}
\begin{figure*}[!htp]
   \begin{center}
   {\includegraphics[width=0.24\linewidth]{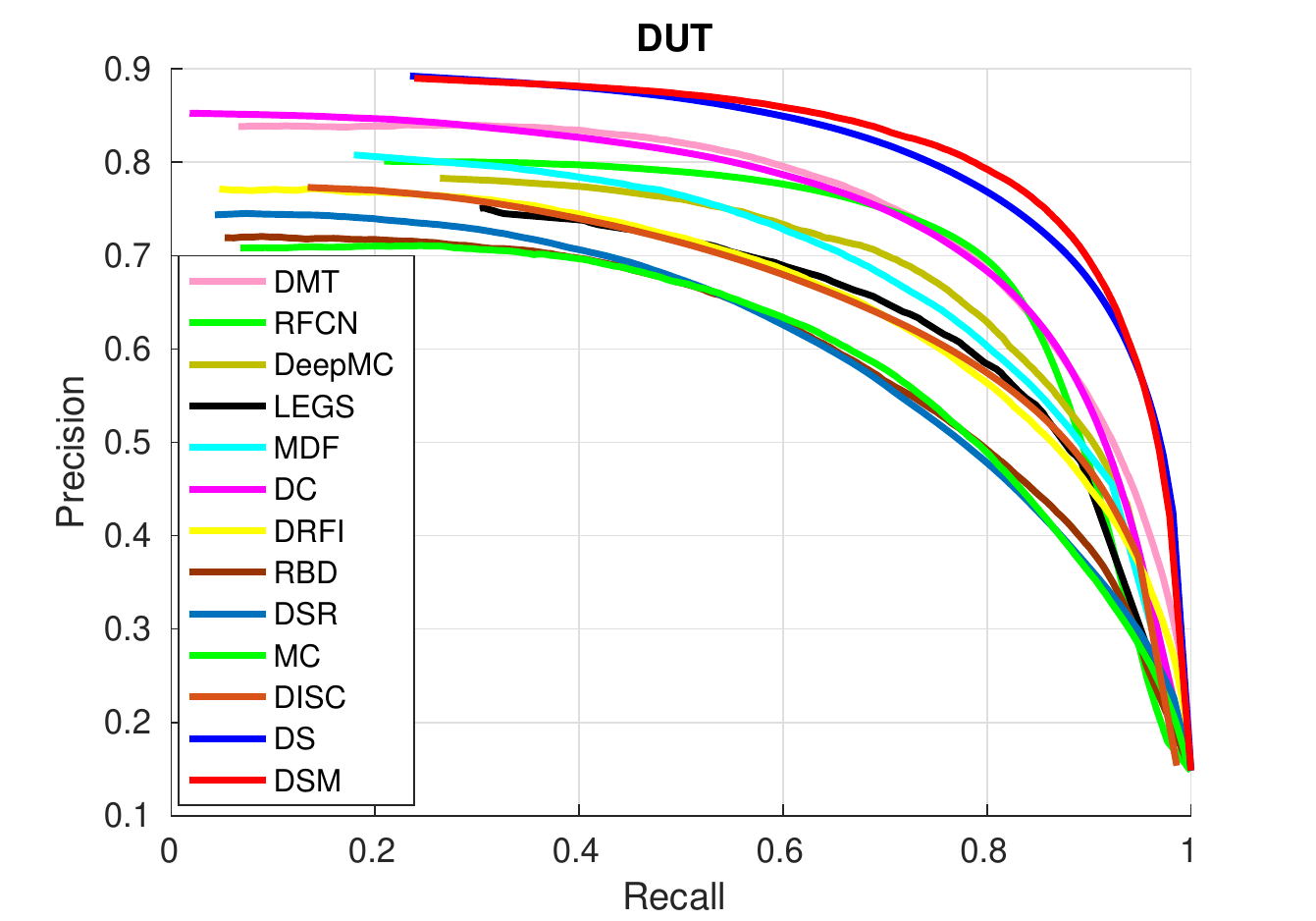}}
   {\includegraphics[width=0.24\linewidth]{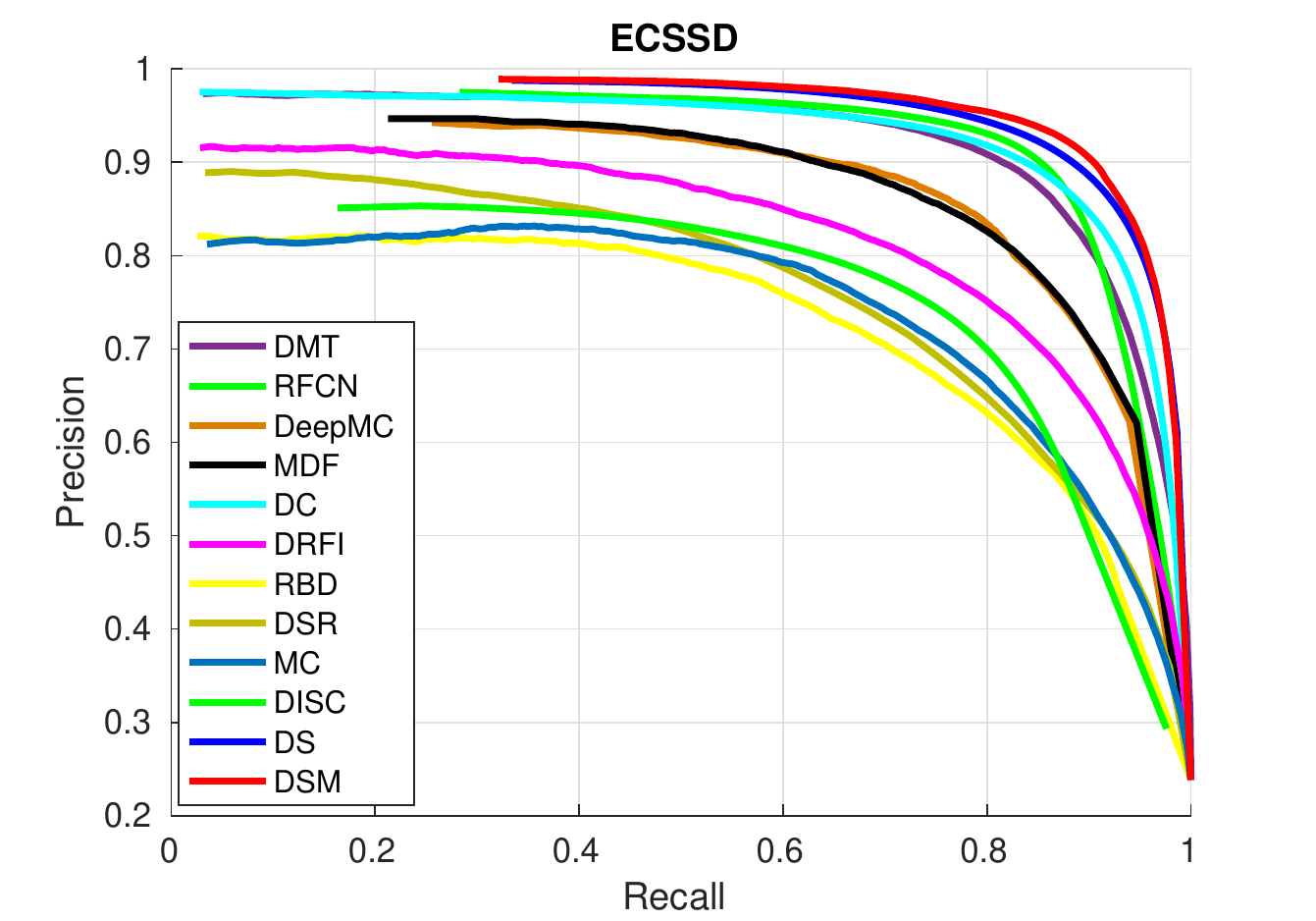}}
   {\includegraphics[width=0.24\linewidth]{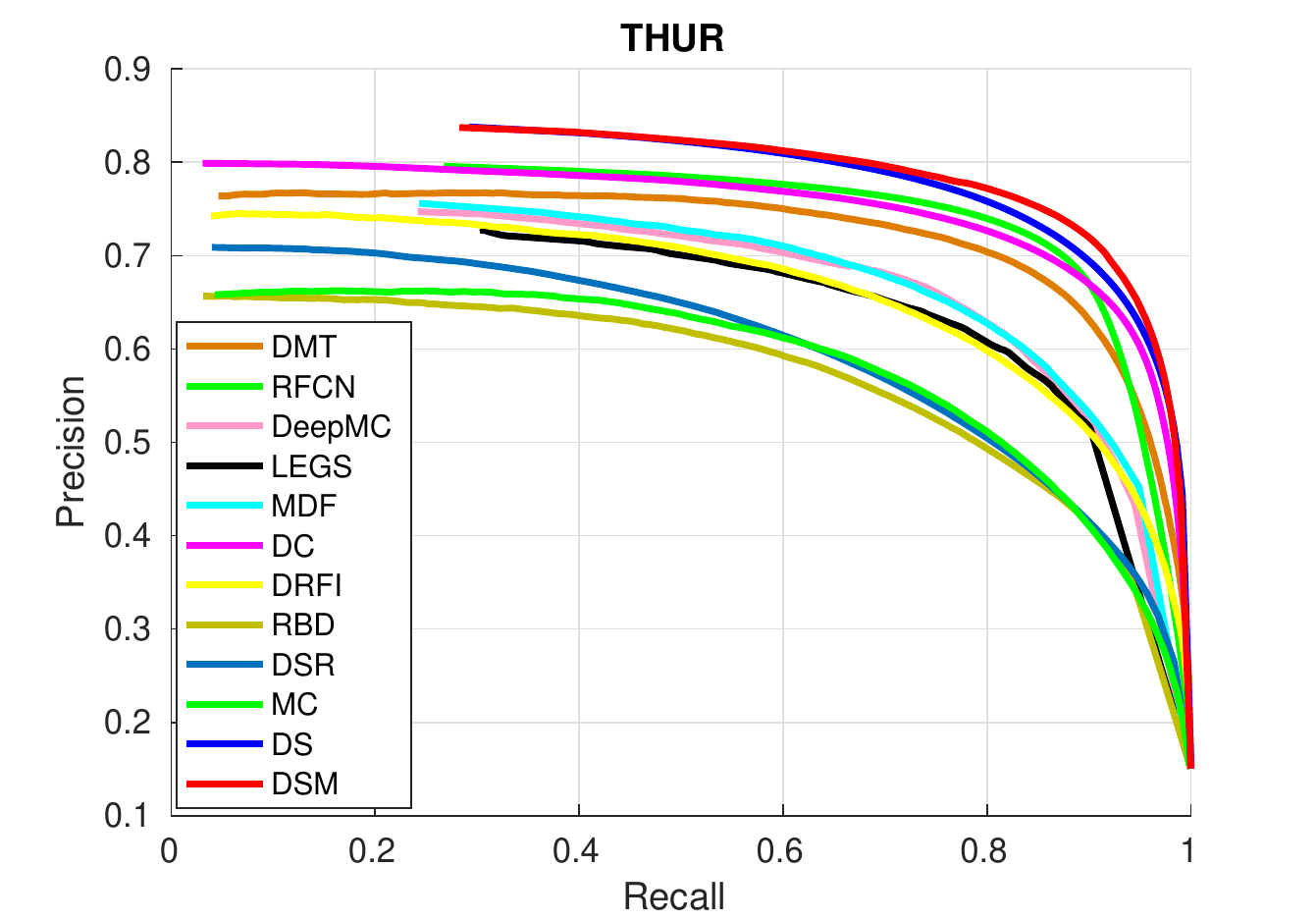}}
   {\includegraphics[width=0.24\linewidth]{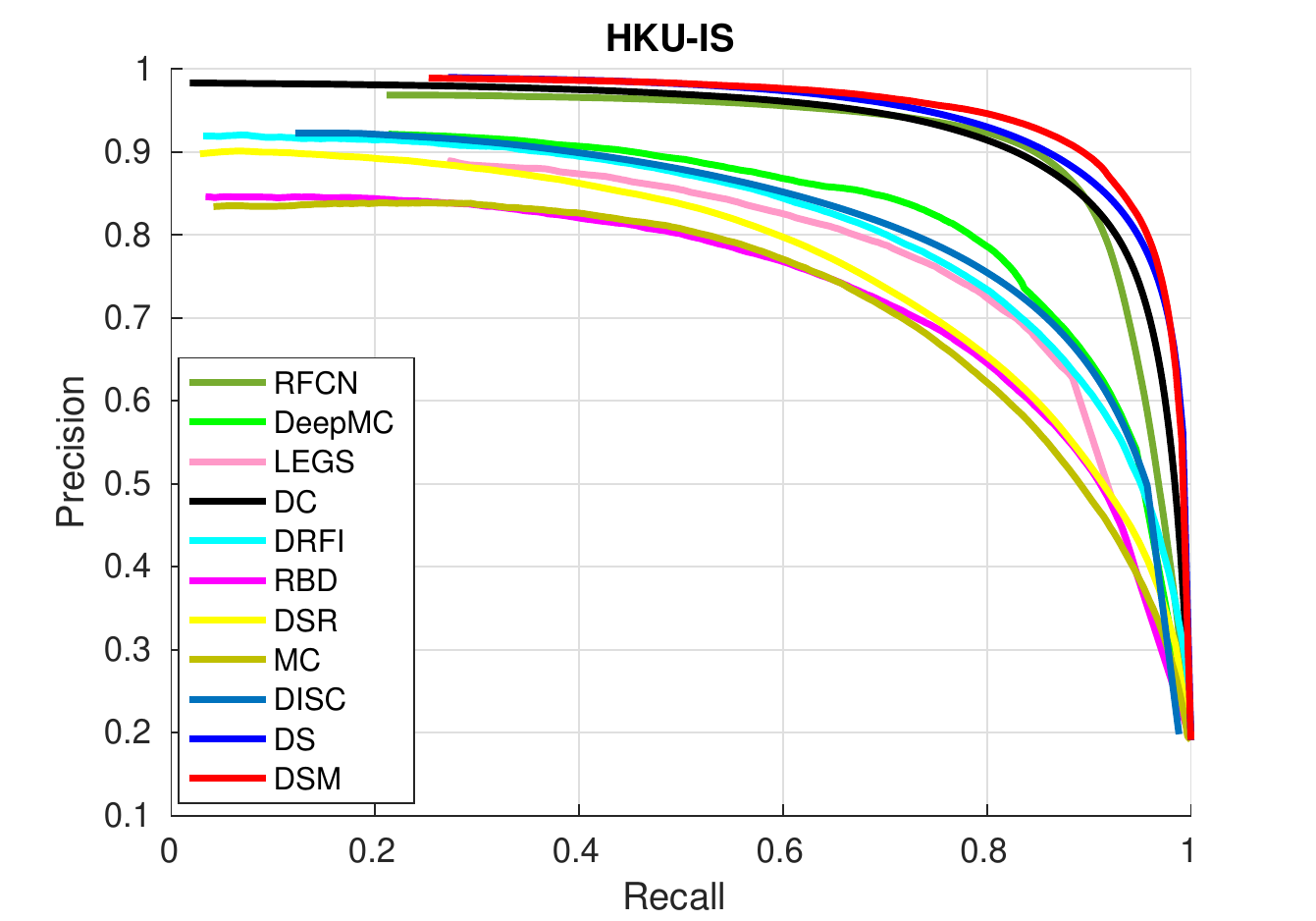}}
   \end{center}
   \vspace{-5mm}
   \caption{Precision-Recall curves on four benchmark datasets (DUT, ECSSD, THUR, HKU-IS). Best Viewed on Screen.}
   \label{fig:pr_curve}
\end{figure*}
\section{Experimental Results}
\begin{figure}[!htp]
   \begin{center}
   {\includegraphics[width=0.9\linewidth]{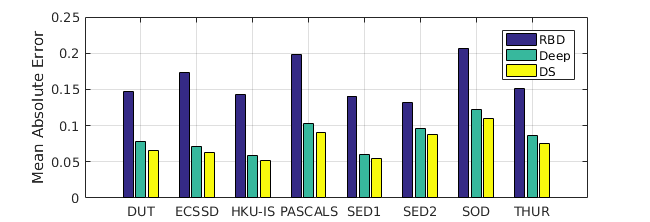}}
   \end{center}
   \vspace{-5mm}
   \caption{MAE on eight benchmark datasets.}
   \label{fig:mae_compare}
\end{figure}

\subsection{Experimental Setup}
\textbf{Dataset:} We trained our deep-shallow model by using 3,000 images from the MSRA10K dataset \cite{ChengPAMI15} for training and 2,000 images for validation. Most of the images in this dataset contains only one salient object. Eight saliency benchmark datasets are used for testing, including ECSSD \cite{Hierarchical:CVPR-2013}, DUT \cite{Manifold-Ranking:CVPR-2013}, SED1 and SED2 \cite{SED}, SOD \cite{DRFI:CVPR-2013}, PASCAL-S \cite{PASCALS}, HKU-IS \cite{MDF:CVPR-2015} and THUR \cite{THUR} dataset.



\textbf{Compared methods:} We compared our method against six state-of-the-art deep learning based methods: DMT \cite{TIP}, RFCN \cite{RFCN}, DISC \cite{DISC}, DeepMC \cite{DeepMC}, MDF \cite{MDF:CVPR-2015} and DC \cite{DC}, and four traditional saliency detection methods: DRFI \cite{DRFI:CVPR-2013}, RBD \cite{Background-Detection:CVPR-2014}, DSR \cite{DSR_ICCV13} and MC \cite{MC_ICCV13} which are proven in \cite{SalObjBenchmark_Tip2015} as the state-of-the-art before the era of deep learning.


\textbf{Training details:}
We trained our model using Caffe \cite{jia2014caffe}, where the training stopped when training accuracy kept unchanged for 200 iterations with maximum iteration 20,000. Each image is scaled to $224\times 224 \times 3$. We initialized our model using Deep Residual Model trained for semantic segmentation \cite{Deeplab}. Weights of the last two convolution layers are initialized using the ``xavier'' policy, bias is initialized as constant. We used stochastic gradient descent method with momentum 0.9 and decreased learning rate 90\% when training loss did not decrease. Base learning rate is initialized as 1e-4 with the ``poly'' decay policy. For loss in both deep model and shallow model, ``Softmaxwithloss'' is utilized. For validation, we set ``test\_iter'' as 1,000 (test batch size 2) to cover the full 2,000 validation images. The whole training takes 58 hours with training batch size 3 and ``iter\_size'' 10 on a PC with an NVIDIA Quadro M4000 GPU.

\textbf{Evaluation metric:} We use two evaluation metrics: mean absolute error (MAE) and Precision-Recall (PR) curve. MAE can provide better estimation of the dissimilarity between the predicted saliency map and the ground truth saliency map, which is defined as:
$ MAE = \frac{1}{W\times H} \sum_{x=1}^W \sum_{y=1}^H |S(x,y)-GT(x,y)|$, where $W$ and $H$ are width and height of the saliency map $S$, $GT$ is the ground truth saliency map. $Precision$ corresponds to the percentage of salient pixels being correctly detected, while $recall$ is the fraction of detected salient pixels in relation to the ground truth number of salient pixels. Precision-Recall (PR) curves are obtained by binarizing the saliency map in the range of [0,255].

\subsection{Model Analysis}
We propose to integrate deep saliency and unsupervised saliency in a unified framework. As shown in Fig.~\ref{fig:saliency_compare1}, there exits scenarios when unsupervised saliency outperform deep saliency a lot. By doing extensive experiments on exiting benchmark datasets, we found out that for some simple scenarios where saliency strongly rely to those basic priors (especially background prior \cite{Background-Detection:CVPR-2014}), unsupervised methods achieves better results than the state-of-the-art deep learning based methods. By utilizing results of unsupervised methods as part of our input, we achieve improved results. To illustrate how unsupervised saliency helps performance of our deep-shallow model, we compute MAE on 8 datasets as shown in Fig.~\ref{fig:mae_compare}, where ``RBD'' represents performance of using RBD \cite{Background-Detection:CVPR-2014}, ``Deep'' represents performance by training our deep model alone, and ``DS'' represents results from our deep-shallow model. As shown in Fig.~\ref{fig:mae_compare}, with the help of unsupervised saliency, our deep-shallow model achieves consistently lower MAE, which can also be observed in Fig.~\ref{how_shallow_helps}.


\subsection{Comparison with State-of-the-art Methods}

\begin{figure}[!htp]
   \begin{center}
   {\includegraphics[width=0.11\linewidth]{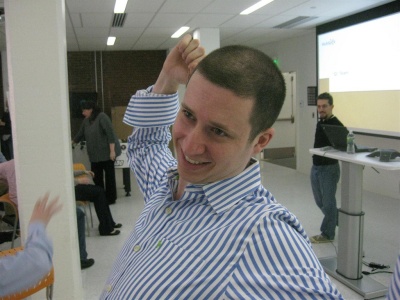}}
   {\includegraphics[width=0.11\linewidth]{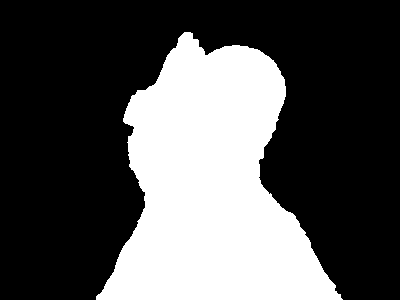}}
   {\includegraphics[width=0.11\linewidth]{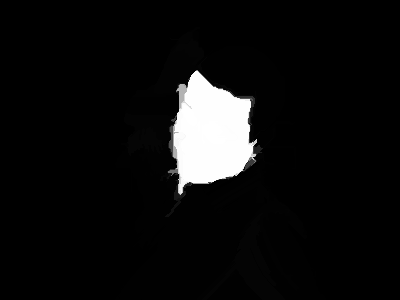}}
   {\includegraphics[width=0.11\linewidth]{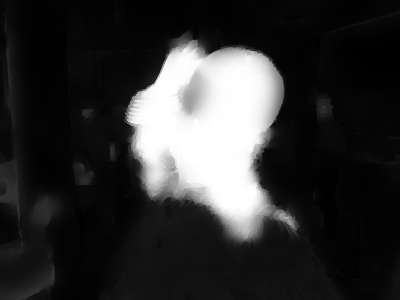}}
   {\includegraphics[width=0.11\linewidth]{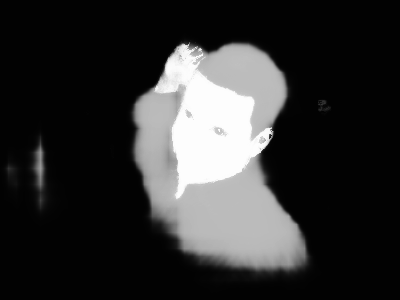}}
   {\includegraphics[width=0.11\linewidth]{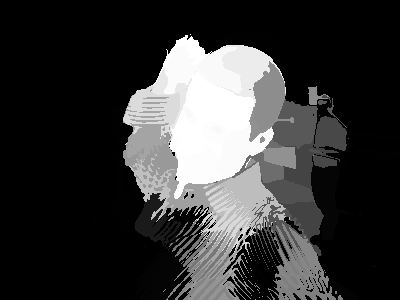}}
   {\includegraphics[width=0.11\linewidth]{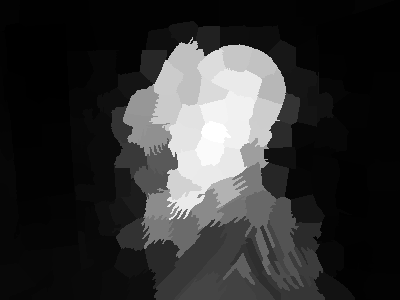}}
   {\includegraphics[width=0.11\linewidth]{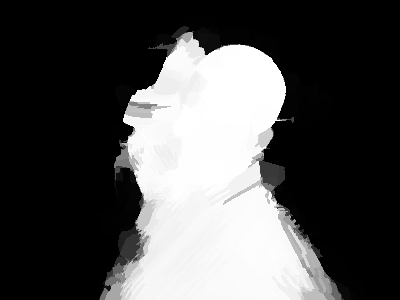}}
   {\includegraphics[width=0.11\linewidth]{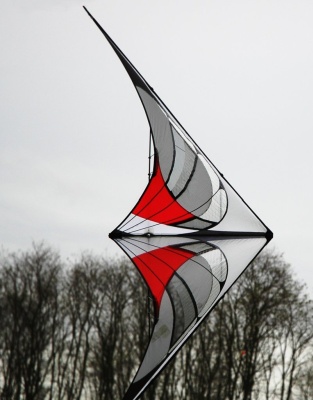}}
   {\includegraphics[width=0.11\linewidth]{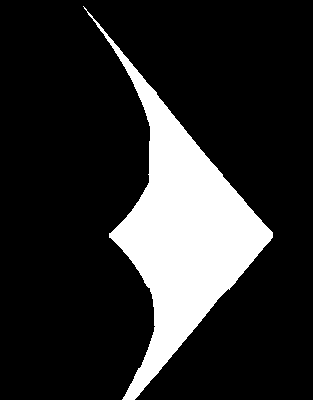}}
   {\includegraphics[width=0.11\linewidth]{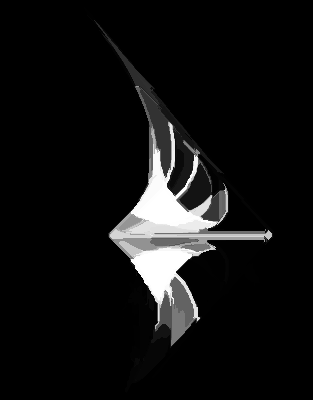}}
   {\includegraphics[width=0.11\linewidth]{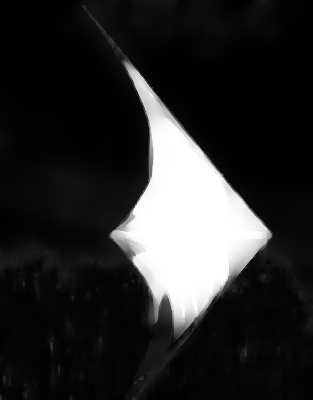}}
   {\includegraphics[width=0.11\linewidth]{compare_fig/0761_dcl_crf.png}}
   {\includegraphics[width=0.11\linewidth]{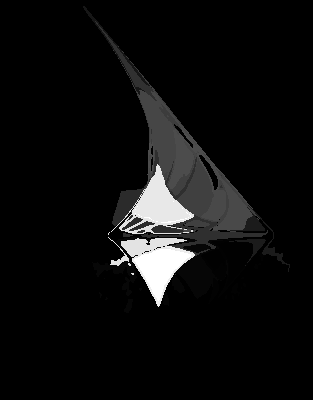}}
   {\includegraphics[width=0.11\linewidth]{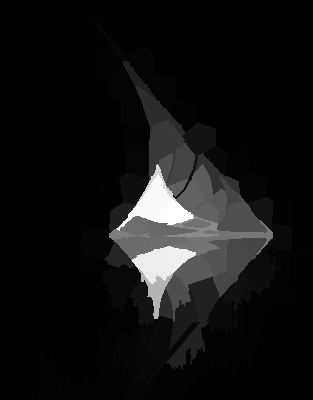}}
   {\includegraphics[width=0.11\linewidth]{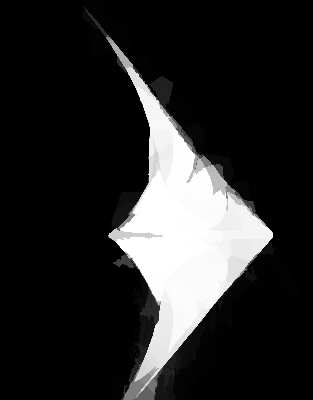}}

   \end{center}
   \vspace{-6mm}
   \caption{Saliency maps generated by different methods for comparison. From left to right: input image, ground truth saliency map, results of MDF \cite{MDF:CVPR-2015}, RFCN \cite{FCN}, DC \cite{DC}, DeepMC \cite{DeepMC}, DMT \cite{TIP} and our method.}
   \label{final_compare}
\end{figure}

We compared our method with six deep learning based methods and four traditional methods. Results are shown in Table \ref{tab:Performance_Comparison}, where ``DS'' represents results from our deep-shallow model, and ``DSM'' represents performance after using MSSF. Table \ref{tab:Performance_Comparison} shows that ``DSM'' achieves consistently smaller MAE compared with ``DS'', which proves the effectiveness of MSSF. Furthermore, for those 8 datasets, deep learning based methods outperform traditional methods with 3\%-9\% decrease in MAE. Our method ``DSM'' achieves consistently the best performance compared with those state-of-the-art methods, especially on the ECSSD, SED1, SED2, PASCALS, SOD and HKU-IS datasets, our method achieves more than 2\% decrease in MAE over the best of the compared methods. In Fig.~\ref{fig:pr_curve}, we show Precision-Recall (PR) curves on four datasets. Our approach consistently outperforms other methods with a wide margin especially on the DUT dataset. Fig.~\ref{final_compare} demonstrates several visual comparisons, where our method outperforms the competing methods.
\begin{figure}[!htp]
   \begin{center}
   {\includegraphics[width=0.18\linewidth]{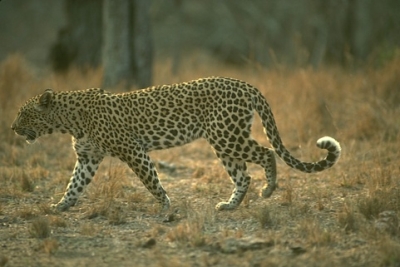}}
   {\includegraphics[width=0.18\linewidth]{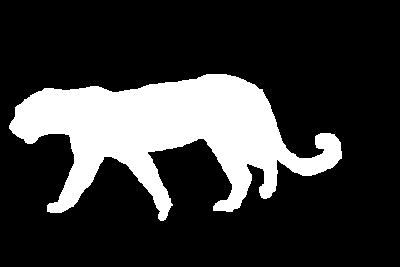}}
   {\includegraphics[width=0.18\linewidth]{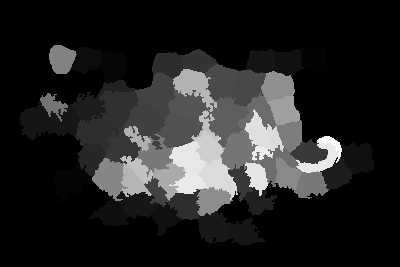}}
   {\includegraphics[width=0.18\linewidth]{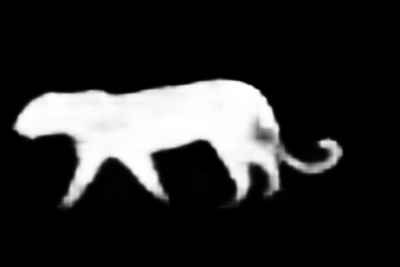}}
   {\includegraphics[width=0.18\linewidth]{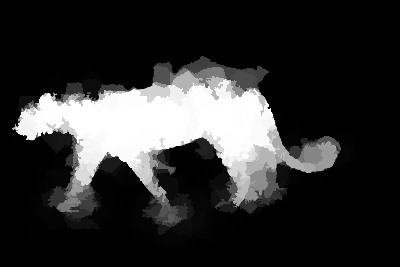}}
   \end{center}
   \vspace{-6mm}
   \caption{Failed example of our model. From left to right: input image, ground truth, results of RBD, deep model and our integrated model.}
   \label{we_fail}
\end{figure}
\section{Conclusions}
By integrating deep saliency and unsupervised saliency for salient object detection, we proposed a unified deep convolutional neural network based method for saliency detection. Our method takes results of unsupervised saliency detection method and the normalized color images as inputs, and directly learns an end-to-end mapping between the inputs to corresponding saliency maps. Then multi-scale superpixel level saliency map fusion is performed to achieve spatially consistent saliency map with sharp object boundaries preserved. Extensive results on 8 benchmark datasets demonstrate effectiveness of our method. While there still exists scenarios when our integrated model failed to improve performance of the deep model as shown in Fig.\ref{we_fail}. and how to get better complementary information of deep feature and low level feature will be our next step.



\bibliographystyle{IEEEbib}
\bibliography{refs}

\end{document}